\documentstyle{article}
\renewcommand{\baselinestretch}{1.1}

\title{\sf Determination of referential property and \\ number of nouns in
Japanese sentences \\ for machine translation into English}
\author{\sf Masaki Murata, Makoto Nagao \\[0.3cm]
\sf Department of Electrical Engineering,\\
\sf Kyoto University, Kyoto, JAPAN\\
\sf \{mmurata,nagao\}@kuee.kyoto-u.ac.jp}
\date{}

\begin{document}
\sf
\maketitle

\begin{abstract}
\sf
When translating Japanese nouns into English, we face the problem of
articles and numbers which the Japanese language does not have, but
which are necessary for the English composition.  To solve this
difficult problem we classified the referential property and the
number of nouns into three types respectively.  This paper shows that
the referential property and the number of nouns in a sentence can be
estimated fairly reliably by the words in the sentence.  Many rules
for the estimation were written in forms similar to rewriting rules in
expert systems.  We obtained the correct recognition scores of 85.5\%
and 89.0\% in the estimation of the referential property and the
number respectively for the sentences which were used for the
construction of our rules.  We tested these rules for some other
texts, and obtained the scores of 68.9\% and 85.6\% respectively.
\thispagestyle{plain}
\end{abstract}

\section{\sf Introduction}
One of the difficult problems in machine translation from Japanese to English
or other European languages is the treatment of articles and numbers.  There
are referential pronominals in Japanese such as KONO, ANO, etc., but these
are used only in particular occasions where references are to be indicated
definitely.  As to the number the Japanese language has no plural form for
nouns and no distinction in verb conjugation to indicate the number of
subject or object of a verb.  In English there are definite and indefinite
articles for nouns and also the distinction between singular and plural.
Therefore the correspondence of articles and numbers for nouns in Japanese
to English translation is a very difficult problem.

To solve this problem to a certain extent, we have to estimate the referential
properties of nouns in a sentential utterance.  It is commonly believed that
the language understanding mechanism is necessary to solve this problem, and
certain contextual or inter-sentential  information is to be grasped.  It is
true, but it is difficult at the present level of natural language analysis
technology.

We propose here that lots of keys exist in the surface information of a
sentence to determine the referential property and the number of a noun in the
sentence.  For example, ``KARE-WA(he)  GAKUSEI(student) DESU(is) '' indicates
that KARE is a specific person(singular), and is linked by a copula to
GAKUSEI, which is a countable noun.  Therefore the property, singular, is
inherited to GAKUSEI from KARE, and the translation is ``He is a student''.
When the above example is changed as ``KARE-WA(he) KINOU(yesterday) ITTO(first)
SHOU-O(prize) MORATTA(was given) GAKUSEI(student) DESU(is)'', where ``student''
is modified by an embedded sentence ``he was given the first prize yesterday'',
this indicates that ``student'' in this sentence is strictly specified, and is
definite.  Therefore the English expression to this Japanese expression is
``He is \underline{the} student who was given the first prize yesterday''.

This sort of judgement is not absolutely reliable but just probable.  This
means that what we have to do is to construct a kind of expert system by
incorporating large number of heuristic rules with certain reliable factors.
In the following we will describe what kind of heuristic rules we have written
for the articulation of the referential property and the number
of a noun in a Japanese sentence.

\section{\sf Categories of Referential Property and Number}

\subsection{\sf Categories of Referential Property}
Referential property of a noun phrase here means
how the noun phrase denotes the subject.
We classified noun phrases into the following three types
from the referential property.
{\scriptsize
\[\rm \mbox{\normalsize \sf noun phrase}
 \left\{ \begin{array}{l}
     \rm \mbox{\normalsize \sf {\bf generic} noun phrase}\\
         \mbox{\normalsize \sf {\bf non generic} noun phrase}
            \left\{ \begin{array}{l}
              \rm \mbox{\normalsize \sf {\bf definite} noun phrase}\\
                  \mbox{\normalsize \sf {\bf indefinite} noun phrase}
            \end{array}
            \right.
\end{array}
\right.
\]
}
A noun phrase is classified as generic
when it denotes all members of the class of the noun phrase
or the class itself of the noun phrase.
For example, ``dogs'' in the following sentence is a generic noun phrase.
\begin{equation}
\mbox{\underline{Dogs} are useful.}
  \label{eqn:doguse}
\end{equation}
A noun phrase is classified as definite
when it denotes a contextually non-ambiguous member
of the class of the noun phrase.
For example, ``the dog'' in the following sentence is a definite noun phrase.
\begin{equation}
\mbox{\underline{The dog} went away.}
  \label{eqn:thedoguse}
\end{equation}
An indefinite noun phrase denotes
an arbitrary member of the class of the noun phrase.
For example, the following ``dogs'' is an indefinite noun phrase.
\begin{equation}
\mbox{There are three \underline{dogs}.}
  \label{eqn:threedogs}
\end{equation}

\subsection{\sf Categories of Number}
Number of a noun phrase is
the number of the subject denoted by the noun phrase.
Categories of number are as follows.
{\scriptsize
\[\rm \mbox{\normalsize \sf noun phrase}
   \left\{\begin{array}{l}
    \rm \mbox{\normalsize \sf {\bf countable} noun phrase}
        \left\{\begin{array}{l}
          \rm\mbox{\normalsize \sf {\bf singular} noun phrase}\\
             \mbox{\normalsize \sf {\bf plural} noun phrase}
          \end{array}
          \right.\\
    \mbox{\normalsize \sf {\bf uncountable} noun phrase}
\end{array}
\right.
\]
}

\section{\sf How to Determinate Referential Property and Number}

\begin{figure}[t]

  \begin{minipage}[c]{14.8cm}
    \begin{center}
      ``KARE(he)-WA SONO(the)-BENGOSHI(lawyer)-NO(of) MUSUKO(son)-NO(of) \\
      HITORI(one person)-DESU(is).'' \, \,
      (He is one of the sons of the lawyer.)

 \center{(a):Japanese sentence}
    \end{center}

  \begin{minipage}[c]{14.8cm}
\hspace*{7.87cm}\protect\verb+KARE(he)-WA----|  +\\
\hspace*{4cm}\protect\verb+SONO(the)----|                      |  +\\
\hspace*{4.37cm}\protect\verb+BENGOSHI(lawyer)-NO(of)----|      |  +\\
\hspace*{7.5cm}\protect\verb+MUSUKO(son)-NO --|  +\\
\hspace*{8.63cm}\protect\verb+HITORI(one person)-DESU(is) +
\center{(b):Dependency structure of sentence(a)}
 \end{minipage}

\vspace{5mm}

 \begin{minipage}[c]{14.8cm}
\small
\baselineskip=12pt

\hspace*{0cm}\protect\verb+( <[noun common-noun +\_\verb+ +\_\verb+ `HITORI'
`HITORI']+\\
\hspace*{0.34cm}\protect\verb+   [copula +\_\verb+ copula DESU-line-basic-form
`DA' `DESU']+\\
\hspace*{0.34cm}\protect\verb+   [punctuation-mark period +\_\verb+ +\_\verb+
`$@!
\hspace*{0.34cm}\protect\verb+   ( <[noun common-noun +\_\verb+ +\_\verb+
`MUSUKO' `MUSUKO']+\\
\hspace*{0.83cm}\protect\verb+      [postpositional-particle
noun-connection-postpositional-particle +\_\verb+ +\_\verb+ `NO' `NO']> +\\
\hspace*{0.83cm}\protect\verb+      ( <[noun common-noun +\_\verb+ +\_\verb+
`BENGOSHI' `BENGOSHI']+\\
\hspace*{1.32cm}\protect\verb+         [postpositional-particle
noun-connection-postpositional-particle +\_\verb+ +\_\verb+ `NO' `NO']> +\\
\hspace*{1.32cm}\protect\verb+         ( <[referential-pronominal   +\_\verb+
+\_\verb+ +\_\verb+ `SONO' `SONO']> )))+\\
\hspace*{0.34cm}\protect\verb+   ( <[noun common-noun +\_\verb+ +\_\verb+
`KARE' `KARE']+\\
\hspace*{0.83cm}\protect\verb+      [postpositional-particle
topic-marking-postposition +\_\verb+ +\_\verb+ `WA' `WA']+\\
\hspace*{0.83cm}\protect\verb+      [punctuation-mark komma +\_\verb+ +\_\verb+
`$@!$(J' `$@!$(J']> ))+
  \end{minipage}
\center{(c):Dependency structure representation of sentence(a)}

\caption{\sf Example of dependency structure
representation}\label{fig:bengoshi_csan}

  \end{minipage}

\end{figure}

\begin{figure}[t]

\fbox{
 \begin{minipage}[c]{14.8cm}
\small
\baselineskip=12pt
\hspace*{0.83cm}\protect\verb+      ( <[noun -] - >+\\
\hspace*{1.32cm}\protect\verb+         ( <[referential-pronominal   +\_\verb+
+\_\verb+ +\_\verb+ `SONO' `SONO']> ) - )+

  \caption{\sf An expression of the noun modified by ``SONO(the)''}
  \label{fig:$@$=$N(J}
  \end{minipage}
}

\end{figure}

\noindent Heuristic rules for the referential property are given in the form:\\
 ({\it condition for rule application}) \\
\hspace*{0.5cm}$\Longrightarrow$ \{
indefinite({\it possibility, value}) \,
definite({\it possibility, value}) \,
generic({\it possibility, value})
\}\\
\noindent Heuristic rules for the number are given in the form:\\
 ({\it condition for rule application})  \\
\hspace*{0.5cm}$\Longrightarrow$ \{
singular({\it possibility, value}) \,
plural({\it possibility, value}) \,
uncountable({\it possibility, value})
\}

\noindent In {\it condition for rule application},
a surface expression is written in the form
like in Figure~\ref{fig:$@$=$N(J}.
{\it Possibility} has value 1 when  the
categories: indefinite, definite, generic, singular, plural or uncountable,
are
possible in the context checked by the condition.
Otherwise the value is 0 for {\it possibility}.
{\it Value} means that
a relative possibility value between 1 and 10
(integer) is given according to
the plausibility of the condition that the {\it possibility} is 1.
Larger value means the plausibility is high.

The rules are all heuristic so that the categories are not exclusive.  In a
certain conditional situation both indefinite and generic are possible, and
also both singular and plural can co-exist.  In these cases, however, the
possibility
values may be different.

Several rules can be applicable to a specific noun in a sentence.
In this case
 the possibility values are added for individual categories
and the final decision of a
category for a noun is done by
the maximum possibility value.
An example is given in Section \ref{subsec:abs_rule}.

When determinating the referential property and the number of nouns,
the condition part is matched not for a word sequence
but for a dependency structure of a sentence.
The dependency structure of a sentence (Figure~\ref{fig:bengoshi_csan}(a)) is
shown in Figure~\ref{fig:bengoshi_csan}(b)
which is represented as Figure~\ref{fig:bengoshi_csan}(c)\footnote{
\sf
This is the result transformed by the system
\cite{csan3}.
    }
to which the condition is checked.
In heuristic rules,
this expression can include a wild card(represented by ``\verb+-+'')
which
can match any partial dependency structure representations.
For example, a noun modified by ``SONO(the)'' is expressed
as in Figure~\ref{fig:$@$=$N(J}.
There are many other expressions
such as regular expressions, AND-, OR-, NOT-operators,
MODee-operator for checking modifyer-modifyee relation and so on.

\vspace{0.3cm}

{\large \sf Algorithm of the Determination of a Category}\\
The following steps are taken for the decision of a category
for the referential property and the number.
\begin{itemize}
\item[(1)]
  Sentences are
  transformed into dependency structure representations.
\item[(2)]
  Dicision is made for each noun from left to right in
  the sentences transformed
  into dependency structure representation.
  This process allows the decision process to make use of
  the referential property and the number already determined
  (see \ref{subsec:abs_rule}(c)(d) for example).
For each noun, the referential property is first determined,
and then the number.
This brings the utilization of referential property of a noun
 when analyzing the number of the noun
(see  \ref{subsec:num_rule}(3) for example).
In these processes
all the applicable rules are used,
{\it possibility} and {\it value} of each category are computed,
and the category for the maximum value is obtained.
An example of the result is shown in Figure~\ref{fig:bengoshi_noun}.
We can also utilize the global information
of a document to which a sentence belongs in
the decision process.
The {\it condition} part, for example,
can check whether there are identical nouns before.
This information is useful for the determination
of the referential property.
\end{itemize}

\begin{figure}[t]
{
 \begin{minipage}[c]{14.8cm}
\small
\baselineskip=12pt
\hspace*{0cm}\protect\verb+( <[noun common-noun +\_\verb+ +\_\verb+ `HITORI'
`HITORI' indefinite singular]+\\
\hspace*{0.34cm}\protect\verb+   [be-verb +\_\verb+ be-verb
DESU-line-basic-form `DA' `DESU']+\\
\hspace*{0.34cm}\protect\verb+   [punctuation-mark period +\_\verb+ +\_\verb+
`$@!
\hspace*{0.34cm}\protect\verb+   ( <[noun common-noun +\_\verb+ +\_\verb+
`MUSUKO' `MUSUKO' definite plural]+\\
\hspace*{0.83cm}\protect\verb+      [postpositional-particle
noun-connection-postpositional-particle +\_\verb+ +\_\verb+ `NO' `NO']> +\\
\hspace*{0.83cm}\protect\verb+      ( <[noun common-noun +\_\verb+ +\_\verb+
`BENGOSHI' `BENGOSHI' definite singular]+\\
\hspace*{1.32cm}\protect\verb+         [postpositional-particle
noun-connection-postpositional-particle +\_\verb+ +\_\verb+ `NO' `NO']> +\\
\hspace*{1.32cm}\protect\verb+         ( <[referential-pronominal   +\_\verb+
+\_\verb+ +\_\verb+ `SONO' `SONO']> )))+\\
\hspace*{0.34cm}\protect\verb+   ( <[noun common-noun +\_\verb+ +\_\verb+
`KARE' `KARE' definite singular]+\\
\hspace*{0.83cm}\protect\verb+      [postpositional-particle
sub-postpositional-particle +\_\verb+ +\_\verb+ `WA' `WA']+\\
\hspace*{0.83cm}\protect\verb+      [punctuation-mark komma +\_\verb+ +\_\verb+
`$@!$(J' `$@!$(J']> ))+
  \end{minipage}
}
\vspace*{-8pt}
\caption{\sf The result of analyzing the sentence in
Figure~1}\label{fig:bengoshi_noun}
\end{figure}

\section{\sf Heuristic Rules}

We have written 86 heuristic rules for the referential property and 48
heuristic rules for the number.  More than half of these rules are just the
implementation of grammatical properties
explained in standard grammar books of Japanese and
English\cite{kanshi}\cite{gentei}\cite{suu}, but
there are many other heuristic rules which we have originally introduced
ourselves.  Some of the rules are given below.

\subsection{\sf Heuristic Rules for Referential
Property}\label{subsec:abs_rule}

\begin{enumerate}
\item When a noun is modified by a referential pronoun, KONO(this), SONO(its),
etc.,\\
then \,
\{
\mbox{indefinite}  (0, 0)\footnote{\sf
(a, b) means the {\it possibility}(a) and the {\it value}(b).
} \,
\mbox{definite}   (1, 2)  \,
\mbox{generic}  (0, 0)
\}\\
Examples: \underline{KONO}(This) \underline{HON-WA}(book)
OMOSHIROI(interesting)\\
\hspace*{3cm} \underline{This book} is interesting.
\item When a noun is accompanied by a particle (WA), and the
predicate has past tense,\\
then \,
\{
\mbox{indefinite} (1, 0) \,
\mbox{definite}   (1, 3) \,
\mbox{generic} (1, 1)
\} \\
Example: \underline{INU-WA}(dog) MUKOUE(away there) {IKIMASHITA}(went)\\
\hspace*{3cm} \underline{The dog} went away.
\item When a noun is accompanied by a particle (WA), and the
predicate has present tense,\\
then \,
\{
\mbox{indefinite} (1, 0) \,
\mbox{definite}   (1, 2) \,
\mbox{generic} (1, 3)
\}\\
Example: \underline{INU-WA} YAKUNITATSU(useful) DOUBUTSU(animal)  DESU(is)\\
\hspace*{3cm} \underline{Dogs}\footnote{
\sf
Both ``a dog'' and ``the dog'' are possible because of the generic subject.
}
 are useful animals.
\item When a noun is accompanied by a particle HE(to), MADE(up to)
or KARA(from),\\
then \,
\{
\mbox{indefinite} (1, 0) \,
\mbox{definite}   (1, 2) \,
\mbox{generic} (1, 0)
\} \\
Example: KARE-O(he) \underline{KUUKOU-MADE}(airport) MUKAE-NI(to meet)
YUKIMASHOO(let us go)\\
\hspace*{3cm} Let us go to meet him at \underline{the airport}.
\end{enumerate}
There are many other expressions which give some clues
for the referential property of nouns, such as
(i) the noun itself,``CHIKYUU(the earth)''[definite],
``UCYUU(the universe)''[definite], etc.,
(ii) nouns modified by a numeral
(Example: KORE-WA(this) ISSATSUNO(one)
\underline{HON-DESU}(book)[indefinite].
(This is \underline{a book}.)),
(iii) the same noun presented previously
(Example: KARE-WA(he) JOUYOUSHA(car)-TO(and) TORAKKU-O(truck)
ICHIDAI-ZUTU(by ones) MOTTEIMASUGA(have),
\underline{JOUYOUSHA}- \\ NIDAKE(car)[definite]
HOKEN-O-KAKETEIMASU(be insured).
(He has a car and a truck, but only the car is insured.)),
(iv) adverb phrases,``ITSUMO(always)'',``NIHON-DEWA(in Japan)'', etc.
(Example: NIHON-DEWA \underline{SHASHOU-WA}(conductor)[generic]
JOUKYAKU(passenger)-NO(of)
KIPPU-O(ticket) SIRABEMASU(check).
(In Japan, \underline{the conductor} checks the tickets of the passengers.)),
(v) verbs, ``SUKI(like)'', ``TANOSHIMU(enjoy)'', etc.
 (Example: WATASHI-WA(I)
\underline{RINGO-GA}(apple)[generic] SUKI-DESU(like).
(I like \underline{apples}.)).

In the case of no clues, ``indefinite'' is given to a noun
as a default value.

\vspace{3mm}

Let us see an example which has several rule applications for the determination
of the referential property of a noun.  KUDAMONO(fruit) in the following
sentence is an example.
\begin{description}
\item [] WAREWARE-GA(We) KINOU(yesterday) TSUMITOTTA(picked)
\underline{KUDAMONO}-\underline{WA} (fruit) AZI-GA(taste) IIDESU(be good).
\end{description}
\hspace*{2cm} \underline{The fruit} that we picked yesterday tastes delicious.

Seven rules are applied
for the determination of the definiteness of this noun.
These are the followings.

\begin{itemize}
\item[(i)] When a noun is accompanied by WA, and the corresponding predicate
has no past tense \\
(KUDAMONO-\underline{WA} AZI-GA \underline{IIDESU}), \\
then
\{
\mbox{indefinite}  (1, 0) \,
\mbox{definite}  (1, 2)   \,
\mbox{generic}  (1, 3)
\}

\item[(ii)] When a noun is modified by an embedded sentence which has the past
tense (TSUMITOTTA),\\
then \,
\{
\mbox{indefinite}  (1,  0) \,
\mbox{definite}    (1,  1) \,
\mbox{generic}  (1,  0)
\}

\item[(iii)] When a noun is modified by an embedded sentence which has a
definite noun accompanied by WA or GA (WAREWARE-GA),
 then \,
\{
\mbox{indefinite}  (1,  0) \,
\mbox{definite}  (1,  1) \,
\mbox{generic}  (1,  0)
\}

\item[(iv)] When a noun is modified by an embedded sentence which has a
definite noun accompanied by a particle (WAREWARE-GA),
 then \,
\{
\mbox{indefinite}  (1,  0) \,
\mbox{definite}  (1,  1) \,
\mbox{generic}  (1,  0)
\}

\item[(v)] When a noun is modified by a phrase which has a pronoun
(WAREWARE-GA),\\
then \,
\{
\mbox{indefinite}  (1,  0) \,
\mbox{definite}    (1,  1) \,
\mbox{generic}  (1,  0)
\}

\item[(vi)] When a noun has an adjective as its predicate
(KUDAMONO-WA AZI-GA \underline{IIDESU}),\\
then \,
\{
\mbox{indefinite}  (1,  0) \,
\mbox{definite}    (1,  3) \,
\mbox{generic}  (1,  4)
\}

\item[(vii)] When a noun is a common noun
(KUDAMONO),\\
then \,
\{
\mbox{indefinite}  (1,  1) \,
\mbox{definite}    (1,  0) \,
\mbox{generic}  (1,  0)
\}

\end{itemize}

As the result of the application of all these rules,
we obtained the final score of
\{
\mbox{indefinite}  (1,  1) \,
\mbox{definite}    (1,  9) \,
\mbox{generic}  (1,  7)
\}
for KUDAMONO,
and ``definite'' is given as the decision.

\subsection{\sf Heuristic Rules for Number}\label{subsec:num_rule}

\begin{enumerate}
\item When a noun is modified by SONO(its), ANO(that), KONO(this),\\
then \,
\{
\mbox{singular}  (1,  3) \,
\mbox{plural}    (1,  0) \,
\mbox{uncountable}  (1,  1)
\}\\
Example: \underline{ANO}(that) \underline{HON-O} (book) KUDASAI (give me)\\
\hspace*{3cm} Give me \underline{that book}.
\item
When a noun is accompanied by a particle WA, GA, MO, O, and there is a numeral
x which modifies the predicate of a sentence, and\\
if \, x = 1 , \, then \,
\{
\mbox{singular}  (1,  2) \,
\mbox{plural}    (1, 0) \,
\mbox{uncountable}  (1,  0)
\} \\
if \, x $\geq$ 2 , \, then \,
\{
\mbox{singular}  (1,  0) \,
\mbox{plural}    (1,  2) \,
\mbox{uncountable}  (1,  0)
\} \\
Example: \underline{RINGO-O}(apple) NIKO(two) TABERU(eat)\\
\hspace*{3cm} I eat two \underline{apples}.
\item
When a predicate, SUKI(like), TANOSHIMU(enjoy), etc. has a generic noun as an
object, and the noun is accompanied by GA(for SUKI), or O(for TANOSHIMU),\\
then
\{
\mbox{singular}  (1,  0) \,
\mbox{plural}    (1,  2) \,
\mbox{uncountable}  (1,  0)
\}\\
Example: WATASHI-WA(I) \underline{RINGO-GA}(apple) SUKI-DESU(like)\\
\hspace*{3cm} I like \underline{apples}.

\end{enumerate}

There are many other expressions
which determine the number of a noun,
such as (i) nouns modified by a numeral
(Example: KORE-WA(this) ISSATSUNO(one) \underline{HON-DESU}(book)[singular].
(This is \underline{a book}.)),
(ii) verbs such as ATSUMERU(collect), AFURERU(be full with),
(Example: WATASHI-WA(I) NEKO-NO(about cat) \underline{HON-O}(book)[plural]
 ATSUMETEIMASU(collect). (I collect \underline{books} on cats.))
(iii) adverbs such as NANDO-DEMO(as many times as ...),
IKURA-DEMO(as much ...)
(Example: RIYUU-WA(reason)[plural] IKURA-DEMO(as much ...) SIMESEMASU(give).
(I can give you a number of reasons.)).

In the case of no clues, ``singular'' is given as a default value.

\section{\sf Experiments and Results}

Experiments of the determination of the referential property
and the number were
done for the following three texts: typical example sentences
in a grammar book ``Usage of the English Articles''\cite{kanshi},
the complete text of a Japanese popular folktale
``The Old Man with a Wen''\cite{kobu},
 a small fragment of an essay ``TENSEI JINGO''.
The rules were written by referring to these sentences
which have good established English translations.
These sentences can be regarded as a training set.
The results of the experiments are shown in Table \ref{tab:learning}.
Here ``correct'' means that
the result was correct.
``Reasonable'' means
that the result is given, for example,
as non-generic but the correct answer was definite, and so on.
``Partially correct'' means that the result was
included in the correct answer.
``Undecidable'' means that we could not judge
which category is correct by our linguistic intuitions.
We obtained 85.5\% success rate for the determination
of the referential properties
and 89.0\% success rate for the numbers for all these learning samples.
The scores of these tables show that the heuristic rules are well adjusted to
these sentences,
and are effective.

To testify the goodness of the rules
we applied these heuristic rules
to the following three other texts: a Japanese popular folktale
``TURU NO ONGAESHI''\cite{kobu},
 three small fragments of an essay ``TENSEI JINGO'',
 ``Pacific Asia in the Post-Cold-War World''
(A Quarterly Publication of The International House of Japan Vol.12,
No.2 Spring 1992).
These test samples have good English translations.
We used them to check the correctness of the results.
The results are shown in Table \ref{tab:test}.
The success rates for the referential property and the number decreased
down to 68.9\% and 85.6\% respectively by these test samples.
These scores show, however, that the rules are still effective.

The success ratio will decrease greatly for the text areas
which handle abstract notions such as philosophy and polytics.
We may have to change and increase heuristic rules for these text areas.
At this moment we cannot say anything about
whether we can write proper heuristic rules for
such complex situations where delicate abstract notions are handled and the
denotation is ambiguous.

As a conclusion we can say the following,
There are of course many expressions and situations
where inter-sentential information is necessary, but without
utilizing it we can achieve a proper guess about the referential property
and the number to a certain extent.
By incorporating this mechanism into a
machine translation system from Japanese into English we will be able to obtain
better translation quality.

\clearpage

\begin{table}

\begin{center}

\renewcommand{\baselinestretch}{1.1}

{\small \sf
\caption{\sf Learning sample}\label{tab:learning}

\begin{tabular}[c]{|l||r|r|r|r||r|||r|r|r|r||r|} \hline
 & \multicolumn{5}{|c|||}{Referential property}  &  \multicolumn{5}{c|}{Number}
 \\ \hline
\multicolumn{1}{|c||}{value}  &  indef  &  def &  gener &  other & total &
singl   &  plural    &  uncount &  other &   total  \\ \hline
\multicolumn{11}{|c|}{Usage of the English Articles(140 sentences, 380 nouns)}
\\ \hline
   correct   &      96  &     184  &      58  &       1  &     339  &    274 &
  32 &      18  &      25  &  349   \\
reasonable&       0  &       3  &       1  &       0  &       4  &       1  &
    1  &       1  &       0  &       3  \\
  partially correct  &       0  &       0  &       0  &       0  &       0  &
    0  &       0  &       0  &      11  &      11  \\
  incorrect  &       4  &      25  &       7  &       1  &      37  &       3
&      10  &       0  &       4  &      17   \\
\hline
\% of correct&    96.0  &    86.8  &    87.9  &    50.0  &    89.2  &    98.6
&    74.4  &    94.7  &    62.5  &    91.8    \\ \hline
\multicolumn{11}{|c|}{The Old Man with a Wen(104 sentences, 267 nouns)} \\
\hline
   correct   &      73  &     140 &       6  &       1  &    222  &    205  &
  24  &       5  &       0  &   234   \\
reasonable&       3  &       4  &       0  &       0  &       7 &       2  &
   0  &       0  &       0  &       2    \\
  partially correct  &       0  &       0  &       0  &       0  &       0   &
     0  &       0  &       0  &       7  &       7  \\
  incorrect  &      11  &      23  &       4  &       0  &      38    &       1
 &      22  &       1  &       0  &      24   \\
\hline
\% of correct&   83.9  &   84.0  &   60.0  &  100.0  &   83.2  &   98.7  &
52.2  &   83.3  &    0.0  &   87.6    \\ \hline
\multicolumn{11}{|c|}{an essay ``TENSEI JINGO''(23 sentences, 98 nouns)} \\
\hline
   correct   &      25  &     35 &      16  &       0  &    76  &      64  &
13 &       0  &       3  &   80    \\
reasonable&       0  &       4  &       2  &       0  &       6  &       2  &
    1  &       0  &       0  &       3   \\
  partially correct  &       0  &       0  &       0  &       0  &       0   &
     0  &       0  &       0  &       6  &       6  \\
  incorrect  &       5  &      10  &       1  &       0  &      16   &       1
&       6  &       1  &       1  &       9  \\
\hline
\% of correct&    83.3  &    71.4  &    84.2  &    -----  &    77.6   &    95.5
 &    65.0  &     0.0  &    30.0  &    81.6   \\ \hline \hline
 & & & & & & & & & &\\[-0.5cm]
average  & & & & & & & & & &\\[-0.1cm]
\hspace*{3mm}\% of appearance &  29.1  &  57.7   &  12.8   &  0.4  &  100.0 &
74.2  &  14.6   &  3.5  &  7.7   & 100.0  \\[-0.1cm]

\hspace*{3mm}\% of correct &  89.4  &  84.0   &  84.2   &  66.7   &  85.5  &
98.2  &  63.3   &  88.5  &  49.1   &  89.0   \\ \hline
\end{tabular}
}
\end{center}

\begin{center}

\renewcommand{\baselinestretch}{1.1}

{\small \sf

\caption{\sf Test sample}\label{tab:test}

\begin{tabular}[c]{|l||r|r|r|r||r|||r|r|r|r||r|} \hline
 & \multicolumn{5}{|c|||}{Referential property}  &  \multicolumn{5}{c|}{Number}
 \\ \hline
\multicolumn{1}{|c||}{value}  &  indef  &  def    &  gener &  other &   total
&  singl  &  plural    &  uncount &  other &   total    \\ \hline
\multicolumn{11}{|c|}{a folktale ``TURU NO ONGAESHI''(263 sentences, 699
nouns)} \\ \hline
   correct   &     109  &    363  &      13  &      10  &   495   &     610  &
    13  &       1  &       1  &     625  \\
reasonable&       6  &      25  &       0  &       0  &      31  &      12  &
    2  &       0  &       0  &      14   \\
  partially correct  &       0  &       0  &       0  &       0  &       0  &
    0  &       0  &       0  &       1  &       1   \\
  incorrect  &      32  &     135  &       6  &       0  &     173   &       2
&      20  &      37  &       0  &      59  \\
\hline
  \% of correct   &    74.2  &    69.4  &    68.4  &   100.0  &    70.8   &
97.8  &    37.1  &     2.6  &    50.0  &    89.4    \\ \hline
\multicolumn{11}{|c|}{an essay ``TENSEI JINGO''(75 sentences, 283 nouns)} \\
\hline
   correct   &      75  &    81  &      16  &       0  &   172  &     197  &
  13  &       2  &       3  &     215   \\
reasonable&       8  &       9  &       1  &       0  &      18 &       3  &
   1  &       0  &       0  &       4    \\
  partially correct  &       0  &       0  &       0  &       0  &       0   &
     0  &       0  &       0  &       3  &       3     \\
  incorrect  &      33  &      51  &       9  &       0  &      93  &       3
&      55  &       3  &       0  &      61    \\
\hline
  \% of correct   &    64.7  &    57.5  &    61.5  &    ----- &    60.8   &
97.0  &    18.8  &    40.0  &    50.0  &    76.0    \\ \hline
\multicolumn{11}{|c|}{Pacific Asia in the Post-Cold-War World(22 sentences, 192
nouns)} \\ \hline
   correct   &      21  &    108  &      11  &       2  &  142 &     157  &
  6  &       1  &       1  &     165   \\
reasonable&       6  &       7  &       0  &       0  &      13 &       3  &
   0  &       0  &       0  &       3   \\
  partially correct  &       0  &       0  &       0  &       0  &       0   &
     0  &       0  &       0  &       0  &       0   \\
  incorrect  &      11  &      24  &       2  &       0  &      37  &       3
&      20  &       1  &       0  &      24    \\
\hline
  \% of correct   &    55.3  &    77.7  &    84.6  &   100.0  &    74.0   &
96.3  &    23.1  &    50.0  &   100.0  &    85.9   \\ \hline \hline
 & & & & & & & & & & \\[-0.5cm]
average  & & & & &  & & & & &\\[-0.1cm]
\hspace*{3mm}\% of appearance&  25.6  &  68.4  &  4.9   &  1.0   &  100.0 &
84.3  &  11.1   &  3.8   &  0.8   &  100.0   \\[-0.1cm]
\hspace*{3mm}\% of correct &  68.1   &  68.7 &  69.0 &  100.0  &  68.9  & 97.4
&  24.6  & 8.9  &  55.6  & 85.6  \\ \hline
\end{tabular}
}
\end{center}
\end{table}

\end{document}